\documentclass{article}

\usepackage{arxiv}
\usepackage{cite}
\usepackage[utf8]{inputenc} %
\usepackage[T1]{fontenc} 
\usepackage[hyphens]{url}  
\usepackage{hyperref}       
\usepackage{booktabs}       
\usepackage{amsfonts} 
\usepackage{amsmath}
\usepackage{amssymb}
\usepackage{nicefrac}       
\usepackage{microtype}      
\usepackage{lipsum}
\usepackage{graphicx}

\def\etal{\emph{et al}.}

\urldef{\githuburl}\url{https://github.com/mattswatson/attack-agnostic-adversarial-attack-detection}

\title{Attack-agnostic Adversarial Detection on Medical Data Using Explainable Machine Learning}

\author{
 Matthew Watson \\
 Department of Computer Science \\
 Durham University\\
 Durham, UK \\
 \texttt{matthew.s.watson@durham.ac.uk}
 \And
 Noura Al Moubayed \\
 Department of Computer Science \\
 Durham University\\
 Durham, UK \\
 \texttt{noura.al-moubayed@durham.ac.uk}
 \And
}

\begin{document}
\maketitle
\begin{abstract}
Explainable machine learning has become increasingly prevalent, especially in healthcare where explainable models are vital for ethical and trusted automated decision making. Work on the susceptibility of deep learning models to adversarial attacks has shown the ease of designing samples to mislead a model into making incorrect predictions. In this work, we propose a model agnostic explainability-based method for the accurate detection of adversarial samples on two datasets with different complexity and properties: Electronic Health Record (EHR) and chest X-ray (CXR) data. On the MIMIC-III and Henan-Renmin EHR datasets, we report a detection accuracy of $77\%$ against the Longitudinal Adversarial Attack. On the MIMIC-CXR dataset, we achieve an accuracy of $88\%$; significantly improving on the state of the art of adversarial detection in both datasets by over $10\%$ in all settings. We propose an anomaly detection based method using explainability techniques to detect adversarial samples which is able to generalise to different attack methods without a need for retraining.
\end{abstract}


\section{Introduction}
Recently, applications of machine learning in healthcare have shown great success. Machine learning models trained on EHR data are able to predict (with high accuracy) heart failure \cite{choi2016retain}, interpret mammograms \cite{wu2019deep} and diagnose CXR \cite{majkowska2020chest}, and in some cases can match the performance of human experts. However, it is now well demonstrated that such models are susceptible to adversarial attacks: attacks that generate samples designed to mislead a machine learning model into making an incorrect prediction \cite{goodfellow2015explaining}. Examples of such attacks are also effective on medical data such as EHR \cite{an2019longitudinal} and medical imaging data \cite{finlayson2018adversarial}. The presence of adversarial attacks is of particular concern in the medical domain as it would be unethical to deploy a machine learning model to clinical practice if it is considered vulnerable to such malicious attacks, even if the likelihood of an attack is low \cite{finlayson2018adversarial}.

Healthcare ML models are at particular risk of adversarial attacks \cite{finlayson2018adversarial, ma2019understanding, kelly2019key}. Fraud is already pervasive in the US' healthcare economy, with institutions systematically inflating costs and physicians billing for the largest amount possible \cite{finlayson2018adversarial, kalb1999health} and, with machine learning algorithms l ikely to be used for medical decisions in the near future \cite{topol2019high}, adversarial attacks on ML models will be a new avenue for fraud to occur. The pharmaceutical and medical device markets are also domains where adversarial attacks on medical machine learning systems are a risk. The large amounts of money involved in these markets (the median revenue for a single cancer drug is estimated to be \$1.67 billion \cite{prasad2017research}) combined with the increasing number of drug/device approval decisions being made based on digital surrogates for patient responses (for example, in medical imaging \cite{pien2005using}) means that extremely valuable decisions are being made by machine learning algorithms and as such are a likely target for adversarial attacks.

There are also technical vulnerabilities present in many ML models used in healthcare \cite{ma2019understanding, lu2017no}: from low variance in training sets to similar models being used for many different tasks increasing their vulnerability to attacks. Healthcare professionals commonly cite susceptibility to adversarial attacks as a challenge to further adoption of ML in healthcare \cite{kelly2019key}, with the UK's National Health Service (NHS) identifying it as a problem that must be overcome for a machine learning model to be used within the healthcare system \cite{morley2019nhsx}. For these reasons, it is prudent to develop methods of defending against, and detecting, adversarial attacks to provide trust in machine learning solution in medical settings \cite{morley2019nhsx, kelly2019key, ma2019understanding, finlayson2018adversarial}.

In parallel, there has recently been an increased effort to improve the explainability of machine learning models. This area of research aims at explaining the decisions made by black-box machine learning models by making the decisions and the processes behind those decisions understandable to a non machine learning expert \cite{doshivelez2017rigorous}. This has resulted in a number of methods being developed that allow for post- and ante-hoc explanations of models and their decisions \cite{dosilovic2018explainable}.

As adversarial attacks change parts of the input, we hypothesise that ML models place more importance upon these perturbed sections of the input when passed an adversarially perturbed sample. This paper introduces a method that utilises techniques from explainable ML to detect when an adversarial sample is passed to a model by inspecting the areas of the input that the model deems most important. The paper's main contributions are: \textbf{I)} The first adversarial sample detection technique that works effectively with EHR data. \textbf{II)} We propose a novel and simple method for detecting adversarial attacks using explainable techniques and demonstrate that it beats the state of the art on both medical imaging and EHR data despite the sparse, temporal and high-dimensional nature of the data. \textbf{III} The method is model agnostic and will support any machine learning model \textbf{IV)} By framing the adversarial detection as an anomaly detection problem this work presents an approach that generalises to any attack type without the need to retrain \footnote{Supporting code can be found at: \githuburl}.

\section{Related Work}

In this section we provide an overview of current state-of-the-art in explainability and techniques for adversarial generation and detection.

\subsection{Adversarial Attacks on Medical Data}

Adversarial attacks have been developed for numerous data modalities and scenarios \cite{qiu2019review}. Finlayson \etal \cite{finlayson2018adversarial} demonstrate that, despite the challenges that medical imaging data presents, traditional adversarial attack techniques such as Projected Gradient Descent (PGD) \cite{madry2017towards} and patch attacks \cite{brown2017adversarial} can still successfully produce inputs that force a classifier to predict the incorrect label.

The authors in \cite{carlini2017towards} demonstrated that adversarial samples can be transferable across models. The authors also introduced a set of three attacks, known as C\&W attacks, that are capable of bypassing some of the most robust machine learning models to adversarial attacks.

Longitudinal AdVersarial Attack (LAVA) \cite{an2019longitudinal} is designed to generate attacks that are effective on EHR data. LAVA is a saliency score based method that works on discrete and sequential EHR data. By utilising the saliency score it avoids perturbing features that would easily be detected by a human expert whilst maintaining a minimal number of perturbations. The authors showed that it can reduce the accuracy of an attention-based model from $50\%$ to $8\%$. 

\subsection{Adversarial Attack Detection}

Metzen \etal{} \cite{metzen2017detecting} show that it is possible to detect adversarial samples, despite their imperceptible feature changes, through the training of a simple binary classifier. Feinman \etal{} \cite{feinman2017detecting} proposed methods of detecting adversarial samples utilising density estimates of the final hidden layer of the model, and a Bayesian uncertainty estimate. These are designed to complement each other: the density estimate detects adversarial samples as they tend to lie outside the data manifold, and the Bayesian uncertainty detects points in low-confidence regions of the input space. Ma \etal{} \cite{ma2019understanding} show that the methods of \cite{feinman2017detecting} can also be successfully applied to medical imaging data. However, no adversarial detection methods have yet been proposed to work on EHR data mainly due to the challenge of dealing with its temporal dependency and high-dimensionality. Significantly, these adversarial detection methods are extremely model-dependent (e.g. Bayesian uncertainty requires dropout networks) and most require retraining for different types of adversarial attacks.

The methods presented in \cite{lee2019simple} are designed to detect any abnormal sample that is sufficiently far away from training distribution; this includes adversarial samples, but also out-of-distribution samples. The method is based on the probability density of the test sample on the feature space of the neural network using a generative classifier and is able to generalise to unseen attack methods with only a small reduction in accuracy. However, it is only able to be used with classifiers which utilise Softmax.

ML-LOO \cite{yang2019detecting} uses the Leave-One-Out (LOO) explainability technique to detect adversarial samples. LOO is a feature attribution method that uses the reduction in the probability of the selected class when the feature is masked/removed. The authors show that LOO results in the best performance for adversarial detection when compared with other feature attribution methods, however it can be very computationally expensive to compute and as such is impractical when datasets contain a large number of features (i.e. CXR images).

\subsection{Explainable Machine Learning}

The development of explainable machine learning techniques has significantly increased recently. This is mainly driven by regulator's and end user's increased awareness of the impact of machine learning models and the need to understand their decisions, for example, the European Union's General Data Protection Regulation adopts a ``right to an explanation"\footnote{\url{https://gdpr-info.eu/recitals/no-71/}}.
This is of particular interest in healthcare, where interpretable machine learning is seen highly important due to the need for ethical and validated decision making \cite{morley2019nhsx}. A comprehensive review of explainable methods can be found in \cite{carvalho2019machine}. The most common method to date is SHAP \cite{lundberg2017unified}. SHAP approximates the change in expected model prediction when conditioning on each (combination of) feature(s) and is closely tied to Shapley values \cite{lipovetsky2001analysis}. Lundberg \etal{} \cite{lundberg2017unified} propose a number of both model-agnostic and model-specific approximations that enable the practical computation of SHAP values.

In contrast, ante-hoc explanation methods focus on building a model with explainability in mind from the start. This includes many traditional machine learning techniques, but also more recent developments such as Generalised Additive Models plus Interactions \cite{lou2013accurate}, which extends the idea of Generalised Additive Models by modelling the interaction between pairs of features alongside the modelling of individual features through splines. It is important to note that there is an inherent trade-off between accuracy and explainability, and one must choose the correct balance for different applications \cite{harder2019interpretable}.

\section{Methodology}

We introduce novel solutions that utilise SHAP values to detect adversarial attacks and demonstrate that it works on both medical imaging and EHR data. The proposed solutions consist of both fully- and semi-supervised methods, and exploits the differences between the distribution of SHAP values of genuine and perturbed samples in order to accurately detect adversarial samples. Furthermore, as SHAP values are consistent across the entire genuine dataset, our semi-supervised solution is able to generalise to adversarial attacks generated by alternative methods without the need for retraining.

\subsection{Datasets and Classification Models}

Due to privacy concerns around healthcare data, there has traditionally been few sufficiently large, open datasets available in the literature. We utilise 2 EHR datasets: MIMIC-III \cite{johnson2016mimiciii} and Henan-Renmin\footnote{\url{http://pinfish.cs.usm.edu/dnn/}}, and 1 medical imaging dataset: MIMIC-CXR \cite{johnson2019mimiccxr}.

MIMIC-III \cite{johnson2016mimiciii} is a large EHR dataset collected from the Beth Israel Deaconess Medical Center in Boston. It contains 53,423 records of adult admissions of 38,597 distinct patients to the Intensive Care Unit (ICU) between 2001 and 2012, in addition to 7870 neonatal cases admitted between 2001 and 2008. On average, each admission contains 4579 charted observations and 380 laboratory measurements. All data was collected during routine clinical care and includes bedside monitoring notes, lab and microbiology test results, diagnosis and procedure codes, and demographic information.

The Henan-Renmin dataset contains records from 110,300 patients, with significantly fewer features; 62 features per patient comprised of basic examinations and clinical tests. The class label for each record is a combination of three possible diagnoses: hypertension, diabetes and/or fatty liver.

RETAIN \cite{choi2016retain} is a state-of-the-art model designed specifically to work with EHR data. The model aims to mimic typical physician practice by inspecting EHR data in reverse-time order, such that more influence is given to more recent visits when making the final classification. In order to provide interpretable results, RETAIN has a two-level neural attention model that first detects key visits and then detects the key diagnoses from these visits.

We train RETAIN on the MIMIC-III dataset \cite{johnson2016mimiciii}. This results in an accuracy of $81\%$ when predicting patient mortality. To ensure that our adversarial attack detection method adapts to different datasets, we also train the RETAIN model on the Henan-Renmin dataset to predict hypertension, with an accuracy of $73\%$. Hypertension is chosen as it is the most prevalent single label, providing mostly balanced classes. The RETAIN model is not as accurate with this dataset compared to MIMIC-III mainly due to lower number of features.

MIMIC-CXR \cite{johnson2019mimiccxr}, also collected from the Beth Israel Deaconess Medical Center, is a database of 377,110 chest x-rays from 227,827 studies collected between 2011 and 2016. Each study has an associated free text summary report by a radiologist. The reports are analysed using a label extraction tool such as CheXpert \cite{irvin2019chexpert} to generate 14 weak labels (of different diagnoses) for the x-ray images. On a stratified test set of 687 manually labelled (by a certified radiologist) images from the MIMIC-CXR dataset, \cite{johnson2019mimiccxr} report that the label extraction method has an accuracy of $95\%$ for the Cardiomegaly label.

First, we run CheXpert on the radiologists' reports to extract the diagnosis, resulting in 14 labels, each of which is classified as either a positive mention, a negative mention or an uncertain mention. Following \cite{irvin2019chexpert} we treat all uncertain labels as positive mentions. We focus on the Cardiomegaly label as this is both a common diagnosis and provides a balance between positive/negative labels with a low number of uncertain mentions.

Furthermore, we ignore any chest x-rays whose reports do not contain any mention of Cardiomegaly. If these were included, it would be difficult to apply a label to them without making further assumptions. This process generates a set of weak labels for each chest x-ray. An image may have multiple labels (e.g. a patient may have both Cardiomegaly and Pneumonia). We fine-tune Densenet-121 \cite{huang2017densely} (pre-trained on ImageNet \cite{deng2009imagenet}) on MIMIC-CXR, based on the method presented by Rajpurkar \etal \cite{rajpurkar2017chexnet}, to predict a diagnosis of Cardiomegaly, achieving an accuracy of 82\%.

\subsection{Adversarial Sample Generation}
\label{sec:sampleGeneration}

We use state-of-the-art adversarial sample generation techniques that are known to be successful on medical data. LAVA \cite{an2019longitudinal} is used for the two EHR datasets. Both RETAIN trained on MIMIC-CXR and RETAIN trained on Henan-Renmin see a significant reduction in accuracy, as shown in Table \ref{tbl:modelAcc}. The reduction in accuracy is similar to that reported in \cite{an2019longitudinal}.

\begin{figure*}[h]
   \begin{center}
    
    \includegraphics[width=0.825\textwidth]{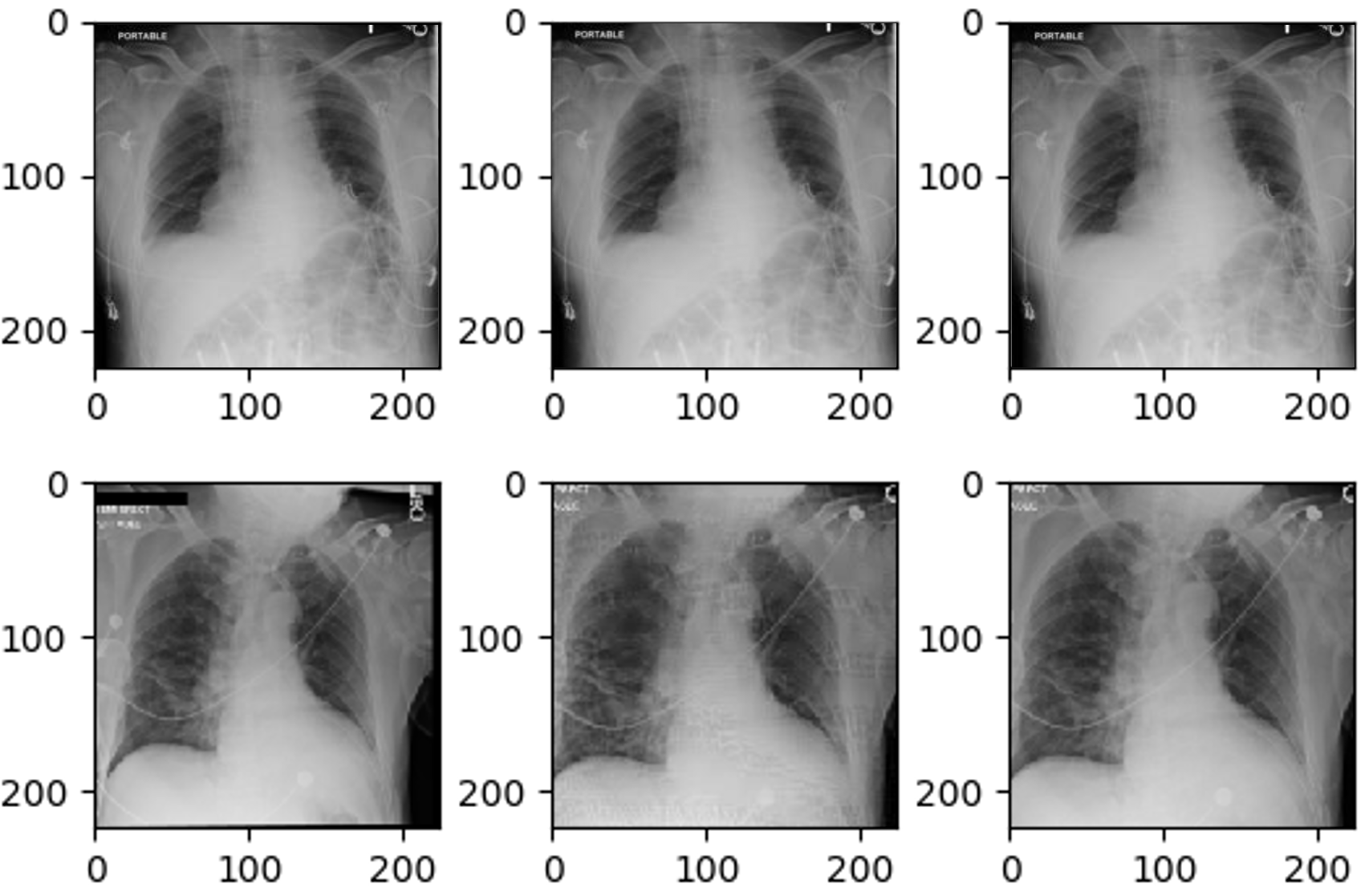}
   
    \caption{Random adversarial examples generated on the MIMIC-CXR. Images on the left are the original images, the middle have been generated via PGD, and the right via C\&W.}
    \label{fig:adv}
     \end{center}
\end{figure*}

\begin{table}[ht]
\centering
\caption{Table showing accuracy of the models on the original and adversarial attack datasets. As PGD necessarily performs perturbations until the sample is classified incorrectly, the MIMIC-CXR model must achieve an accuracy of $0\%$ on the adversarial set.}
\label{tbl:modelAcc}
\begin{tabular}{c|c|c}
\textbf{Model} & \textbf{Acc. original data} & \textbf{Acc. adv. data} \\ \hline
MIMIC-III RETAIN      & $81\%$ & $43\%$ \\ \hline
Henan-Renmin RETAIN   & $73\%$ & $44\%$ \\ \hline
MIMIC-CXR Densenet121 & $82\%$ & $0\%$                                     
\end{tabular}
\end{table}

Projected Gradient Descent (PGD) \cite{madry2017towards} is used to generate the CXR adversarial samples. PGD produces adversarial images to mislead the machine learning model whilst keeping the perturbations small enough that they are not easily detected via traditional methods or even the human eye. Fig. \ref{fig:adv} shows examples generated from random samples in the MIMIC-CXR dataset. As shown in Table \ref{tbl:modelAcc}, PGD successfully produces adversarial samples that are able to mislead the model into making an incorrect classification.

In order to test our method's ability to generalise to different attack types, we use the attack method proposed by Carlini \& Wagner \cite{carlini2017towards} (C\&W). Unlike PGD which uses $L_\infty$ norm, C\&W uses the $L_2$ distance metric to produce a second set of adversarial samples for the MIMIC-CXR dataset. These two approaches are chosen as they perturb the images differently and hence we can better test the generalisation of our approach.

\subsection{Adversarial Attack Classification}
\label{sec:classificationMethod}

\begin{figure*} [h]
    \centering
    \begin{tabular}{cc}
        \includegraphics[width=0.47\textwidth]{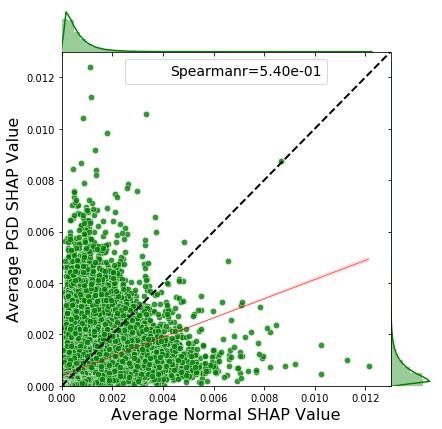}&
        \includegraphics[width=0.47\textwidth]{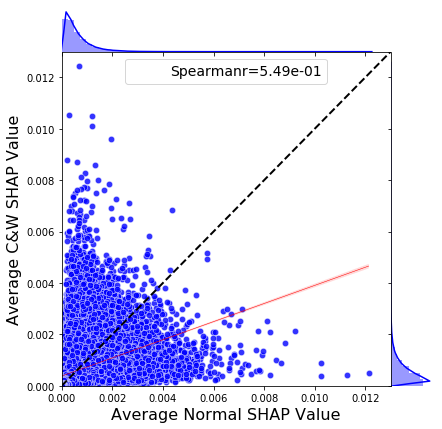}\\
        (a)&(b)
    \end{tabular}
    \caption{Figures showing the average absolute importance of each feature in the original MIMIC-CXR dataset, calculated using SHAP values against the adversarial samples. (a) Scatter plot of the SHAP values of PGD adversarial samples on the Y axis against the SHAP values of original sample on the X axis, the dashed line represents the ideal line while the red line is the linear fit. The histogram of each axis is plotted. The Spearman Rank correlation value is reported.(b) Scatter plot of the SHAP values of C\&W adversarial samples on the Y axis against that of the original set on the X axis.}
    \label{fig:shap}
\end{figure*}

As adversarial attacks subtly change small parts of the input, we hypothesize the SHAP values for an adversarial sample will be different than those for a genuine sample. This is illustrated by Fig. \ref{fig:shap}, which shows how PGD and C\&W affect the distribution of SHAP values compared to the SHAP values of genuine data (correlation is low between the two with most values away from the ideal linear line). This demonstrates that although adversarial attacks methods aim to make the minimal feature perturbations possible, they still greatly impact the distribution of the explanation of the model predictions. Fig. \ref{fig:shap} also demonstrates that the PGD and C\&W attacks perturb the samples differently.

In order to quantify the importance that our models place on different parts of their respective inputs, we utilise SHAP values as calculated by GradientSHAP \cite{lundberg2017unified}. SHAP values reflect the contribution of each individual feature to a model's prediction, which is important when only a small number of features are changed under perturbation during the adversarial attack. 

SHAP values are the (approximate) solution to Eq. \ref{eq:shap}, where $\phi_i(f, x)$ is the importance of feature $i$ of input $x$ to model $f$, $M$ is the number of features, $|z^\prime|$ is the number of non-zero entries in $z^\prime$, $z^\prime \subseteq x^\prime$ represents all $z^\prime$ whose non-zero entries are a subset of the non-zero entries in $x^\prime$ and $S$ is the set of non-zero indices in $z^\prime$.

\begin{equation}
\label{eq:shap}
\begin{split}
    \phi_i(f, x) = \sum_{z^\prime \subseteq x^\prime} \frac{|z^\prime|! (M - |z^\prime| - 1)!}{M!} \big[& \mathbb{E}[f(x) | z_S] - \mathbb{E}[f(x) | z_{S \setminus i}]\big]
\end{split}
\end{equation}

\begin{figure*}[h]
    \centering
    \begin{tabular}{ccc}
        \includegraphics[width=0.29\textwidth]{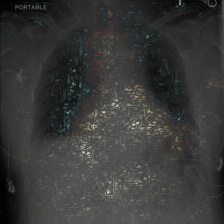}&
        \includegraphics[width=0.29\textwidth]{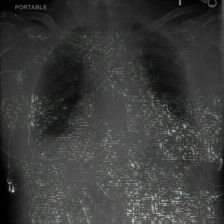}&
        \includegraphics[width=0.29\textwidth]{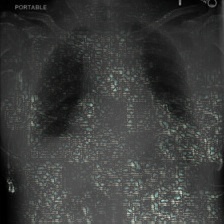}\\
        (a)&(b)&(c)
    \end{tabular}
    \caption{(a) The heatmap of SHAP values overlayed on a genuine sample from the MIMIC-CXR dataset, (b) The heatmap of SHAP values overlayed on the same image after being perturbed via PGD, (c) The heatmap of SHAP values overlayed on the same image after being perturbed by C\&W.}
    \label{fig:heatmapSHAP}
\end{figure*}

We calculate SHAP values for the unperturbed (genuine) dataset and the set of perturbed samples to generate the data for the negative and positive class respectively. Fig. \ref{fig:heatmapSHAP} demonstrates how the SHAP values for a sample change when the model is looking at a perturbed sample, illustrating how a model focuses on different parts of the input when presented with an adversarial sample. The model seems to utilise clusters of pixels in the chest area in the original picture while the important pixels are scatter across the attack images. 

We propose both fully- and semi-supervised methods using SHAP values to detect adversarial samples utilising this information.

\textbf{SHAP-MLP:} We train a simple multi-layer perceptron (SHAP-MLP) on the set of SHAP values from both genuine and adversarial samples of the dataset. The model consists of an input layer, output layer and a single hidden layer. More details about the model are in Section \ref{sec:results}.

\textbf{SHAP-Conv:} We train a convolutional neural network (CNN) on the set of SHAP values from both genuine and adversarial samples. The CNN consists of two convolutional layers, the first going from 3 channels to 16 with a kernel of size 5 and the second going from 16 channels to 32 with a kernel size of 5. We use max pooling with a kernel size and stride of 2, and the ReLU activation function throughout. Following the convolutional layers is a series of 3 fully connected layers of sizes $89888 \times 256$, $256 \times 84$ and $256 \times 1$. We apply dropout with a probability of $0.4$ after the second convolutional layer and again after the second fully connected layer.

\textbf{SHAP-AE \& SHAP-VAE:} Typically, an adversarial attack can be seen as any sample which a model classifies incorrectly; this can include genuine images which the model misclassifies. SHAP-MLP and SHAP-Conv both attempt to classify these images as adversarial. However, it is often more useful to only detect samples which have been specifically perturbed to be adversarial \cite{feinman2017detecting}. This results in a smaller number of samples being present in the adversarial set. Therefore we propose the use of anomaly detection methods to detect the adversarial samples.

We experiment with two semi-supervised models: autoencoders (SHAP-AE) and variational autoencoders (SHAP-VAE) \cite{kingma2014auto} trained to reproduce SHAP values of genuine samples. The reconstruction error of the autoencoder, i.e. the error between the original and reconstructed value, is then used as a measure to detect an adversarial sample. For SHAP-AE, mean squared error (MSE) is used as the loss function. For SHAP-VAE, MSE plus the Kullback-Liebler divergence is used. As the autoencoder is trained only on genuine SHAP values, the reconstruction error from adversarial SHAP values are expected to be higher - the (V)AE has not learned how to reproduce the adversarial values. We thus train an SVM to classify reconstruction error into two classes (adversarial and genuine). The performance of both methods are reported in Section \ref{sec:results}. As both of these methods are semi-supervised approaches, they are able to generalise to different attack types; they learn to reproduce the SHAP values of a genuine dataset, so anything that deviates from that is labelled adversarial. This would be a useful property, as it enables the model to detect novel, unseen attacks.

\section{Experiments and Results}
\label{sec:results}

We report the results of our experiments and compare our approach to the current state of the art adversarial attack detection methods.

\subsection{Experiments on EHR data}
We first report the results of experiments on EHR data. Throughout all experiments, we normalise the SHAP values so they have a mean of 0 and variance of 1, and have a train/test split of 80/20. We train SHAP-MLP on the genuine and adversarial SHAP values from the MIMIC-III dataset. A grid-based cross validation search method is used to find the optimal hyperparameters for SHAP-MLP, resulting in a hidden layer of dimension 160 and a learning rate of 0.01 with the Adam optimiser. This leads to an accuracy of $77\%$. Similarly, on the Henan-Renmin dataset, a hidden layer dimension of 140 and learning rate of 0.01 are optimal, achieving an accuracy of $81\%$.

\begin{table*}[ht]
    \centering
    \caption{Results of adversarial sample detection. HR column reports the accuracy on the Henan-Renmin. CXR (C\&W) reports the accuracy on C\&W generated samples, having been trained on C\&W samples and CXR (PGD) the accuracy of a model trained on PGD samples tested on PGD samples.}
    \label{tbl:experiments}
    \hfill
    
    \resizebox{\textwidth}{!}{
        \begin{tabular}{ |l|c|c|c|c|c|c| }
            \hline
            \textbf{Method} & \multicolumn{6}{|c|}{\textbf{Datasets}}  \\
            \hline
            & MIMIC-III & HR & CXR (C\&W) & CXR (PGD) & CXR (Train: PGD;Test: C\&W) & CXR (Train: C\&W;Test: PGD) \\
            \hline
            SHAP-MLP & \textbf{77\%} & \textbf{81\%} & \textbf{100\%} & 99\% & 58\% & 46\%\\
            SHAP-AE + SVM & 65\% & 53\% & 79\% & 79\%& 77\% & 79\%\\
            SHAP-VAE + SVM & 66\% & 53\% & 85\% & 88\%& \textbf{86\%} & \textbf{88\%} \\
            SHAP-Conv & N/A & N/A & \textbf{100\%} & \textbf{100\%} & 55\% & 65\% \\
            Kernel Density \cite{feinman2017detecting} & 67\% & 67\% & 84\%& 83\% & 72\% & 66\% \\
            ML-LOO \cite{ma2019understanding} & N/A & N/A & 71\% & 78\% & 71\% & 71\%\\
            \hline
        \end{tabular}
    }
\end{table*}

A similar approach is used for testing the autoencoder-based methods. SHAP-AE and SHAP-VAE are both trained on the set of genuine SHAP values from MIMIC-III and Henan-Renmin. After performing the same hyperparameter optimisation method described above, we find that an autoencoder with 2 hidden layers (in both the encoder and decoder), a code size of 20 and a learning rate of 0.01 with an Adam optimiser provides optimal results. Experiments find that an SVM with an RBF kernel with $C=1$ and $\gamma = \frac{1}{M}$ (where $M$ is the number of features) gives the best results compared to logistic regression, and SVMs with other parameters, that are validated using grid-based cross validation search. Similarly, SHAP-VAE has a code size of 5 and a learning rate of 0.01 with an Adam optimiser. For the loss function, the MSE is added to the Kullback-Leibler divergence. An SVM using an RBF kernel with $C=1$ and $\gamma = \frac{1}{M}$ (where $M$ is the number of features) gives the optimal results.

\begin{figure}[h]
    \centering
    \includegraphics[width=7.3cm]{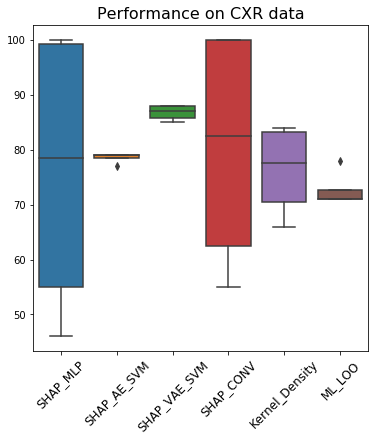}
    \caption{Box plot reporting the performance of adversarial sample detection methods on CXR data.}
    \label{fig:boxplot}
\end{figure}

\subsection{Experiments on Imaging Data}

To test the proposed solutions ability to work on different data modalities, we run the same set of experiments on MIMIC-CXR data. CNNs are shown to achieve superior performance when compared to other model structures \cite{sharma2018analysis}, hence the use of convolutions in SHAP-Conv allows the model to work well on imaging data. This is highlighted by the fact that they outperform all other methods on all medical imaging experiments carried out with a 100\% accuracy on both attack types (Table \ref{tbl:experiments}). Class imbalances in the dataset do not affect our results as our adversarial attack detectors work on balanced classes (non-perturbed images and perturbed images), and we have chosen to focus on the Cardiomegaly label within MIMIC-CXR as it itself provides a balance of positive/negative classes.

To test the semi-supervised models' ability to generalise to different attack types, we test the models trained on the MIMIC-CXR PGD data on MIMIC-CXR data perturbed by the C\&W attack and vice versa. Table \ref{tbl:experiments} shows that both SHAP-AE and SHAP-VAE are able to generalise to different attack types, achieving identical accuracy when C\&W-perturbed examples are added to the test set, confirming that our model can generalise to different attack methods without the need for retraining. This is extremely useful, as it means our model is able to detect unseen attacks. However, as SHAP-MLP and SHAP-Conv are both fully-supervised and are trained on both the genuine and adversarial samples, they are unable to generalise to different attack types. Interestingly, while neither model are able to generalise, SHAP-Conv performs better when trained on PGD images whereas SHAP-MLP achieves a better performance when trained on the C\&W samples. This could indicate that PGD perturbs images in such a way that higher-level features are affected (which will be more difficult for SHAP-MLP to detect), whereas C\&W changes features on a lower level which SHAP-MLP has more success in recognising.

The ability of SHAP-AE and SHAP-VAE (both with SVMs) to generalise to different adversarial attack techniques is further demonstrated through Fig. \ref{fig:boxplot}; both of these techniques have a significantly smaller inter-quartile range than the other techniques tested, showing that the performance of these models is not affected by the type of attack that they are attempting to detect. SHAP-VAE is the clear best performer on CXR data with a stable high performance in all settings.

\subsection{Comparison to existing methods}

The adversarial sample detection method outlined in \cite{ma2019understanding} is used to run the kernel density based adversarial detection method presented in \cite{feinman2017detecting} on the MIMIC-CXR and MIMIC-III datasets. We estimate the kernel density of the final hidden layer of Densenet-121 and RETAIN respectively, performing grid-based cross validation search to find the optimal bandwidths, and fitting a logistic regression classifier on the estimated densities to detect adversarial samples. A bandwidth of 0.1 produces optimal results; the results are reported in Table \ref{tbl:experiments}.  This method is unable to generalise to different attack types without retraining, as the accuracy drops to $66\%$ when C\&W attacks are introduced into the test dataset.

We also compare our methods against the state-of-the-art explainability-based adversarial detection method ML-LOO \cite{yang2019detecting}. We follow the experiments of the authors on Densenet-121, extracting the LOO features from the same layers and utilising the inter-quartile range of these feature attribution maps. We test ML-LOO's ability to generalise in the same way as SHAP-AE and SHAP-VAE. ML-LOO is able to maintain comparable accuracy on the unseen attack type with a $>10\%$ lower detection accuracy compared to SHAP-VAE. The Leave-One-Out (LOO) feature attribution method is also extremely computationally intensive, and is impractical for datasets with large feature spaces. Our method, however, does not suffer from the same issue as we are able utilise one of many possible approximations when calculating SHAP values (for example, throughout this paper the GradientSHAP approximation \cite{lundberg2017unified} is used).

Our proposed methods outperform the state of the art on all data modalities, as reported in Table \ref{tbl:experiments}. Additionally, SHAP-AE and SHAP-VAE are both able to generalise to different attack types without retraining. In contrast, Kernel Density suffers a significant drop in accuracy when tested on unseen attack types in the test set, showing it is unable to accurately classify attacks it has not been trained on, while ML-LOO maintains it is performance but at a significant computational cost. Our results are compatible with those of \cite{ma2019understanding, finlayson2018adversarial} in terms of EHR being a more difficult data to address with SHAP-MLP beating Kernel Density's performance by over $10\%$ in accuracy.

\section{Discussion}

The presented results demonstrate the difficulty to detect adversarial attacks on EHR data. This is due to both the challenges associated with the data, and how LAVA generates adversarial samples; unlike the PGD and C\&W attacks on medical imaging data, LAVA is a saliency-based attack method. This results in smaller changes being made to the SHAP values of adversarial samples, and so they are naturally more difficult to detect.

The MIMIC-CXR data is easier to work with. However, through inspection of the distribution of original labels of the adversarial examples that our model fails to detect, we find that for all labels apart from Cardiomegaly (the label our model is trying to predict) the distribution of positive/negative labels is the same as in the original dataset. However, upon investigation of the distribution of Cardiomegaly labels, we find that our semi-supervised adversarial detection methods incorrectly classifies a higher proportion of positive samples as adversarial than negative samples ($40\%$ of the incorrectly classified samples are CXRs with the Cardiomegaly diagnosis, whereas in the dataset only $29\%$ of images have the label). This shows that class imbalance in the dataset leads to difficult-to-detect adversarial samples. As the original model will most likely have an inherent difficulty to classify one of the classes (due to the class imbalance in the training data), the adversarial sample classifier needs to learn to classify \textit{both} perturbed samples and misclassified-genuine samples as adversarial. As the SHAP values of misclassified-genuine samples will be much closer to that of the genuine training set, this is difficult to do.

The ability of all the proposed models to work on different datatsets is useful in medical scenarios where multi-modal data \cite{xu2019multimodal} and non-standardised data formats \cite{finlayson2018adversarial} are common. Additionally, the ability to detect adversarial samples from unseen adversarial attacks is invaluable, as it reduces the need for bespoke detection techniques to be developed when new attack methods are discovered.

\section{Conclusion}

We present a novel method of detecting adversarial samples using SHAP values that is able to adapt to different attack types and data modalities. Our method is the first such method designed specifically to work on both EHR and medical imaging data, despite the challenges of high-dimensionality, sparsity and temporality that it presents, and as such beats the current state of the art adversarial attack detection techniques on these data modalities. It is also able to generalise to different attack methods without any additional training. By using SHAP values we are able to explain how different attack methods work on different datasets, and use this information to detect samples which have been adversarially perturbed.

Further work will investigate the possibility of modifying current attack methods such as PGD and C\&W to minimise the perturbation of SHAP values rather than features, and explore the effectiveness of such an attack against our detection methods. Additionally, it will explore how explainability, and SHAP in particular, can be used to inspect the distributions of different types of generated data (for example, synthetic data) and utilise these findings to evaluate the usefulness of such data.

\section*{Acknowledgements} 
The authors would like to thank Cievert Ltd and the European Regional Development Fund for sponsoring this work.

\bibliographystyle{acm}
\bibliography{main}

\section{Supplementary Material}

The following figures are randomly sampled images from the MIMIC-CXR dataset. On the left is the original image, in the middle is the same sample, but adversarially perturbed through the PGD attack, and on the right is the sample adversarially perturbed via the C\&W attack. All images are overlayed with the SHAP values of a Densenet-121 model when predicting the presence of Cardiomegaly. These figures show how an adversarial attack significantly changes the SHAP values of a model (the SHAP values for genuine samples are concentrated around the chest area, whereas they are more widely spread for adversarial samples) while keeping the perturbations so small they are imperceptible to the human eye. The figures also demonstrate how the PGD and C\&W attacks produce different SHAP distributions.

\begin{figure*}
    \centering
    \begin{tabular}{ccc}
\includegraphics[width=0.300\textwidth]{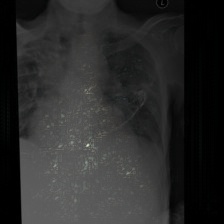}&\includegraphics[width=0.30\textwidth]{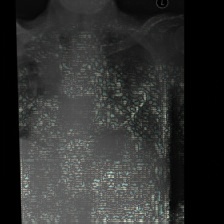}&\includegraphics[width=0.30\textwidth]{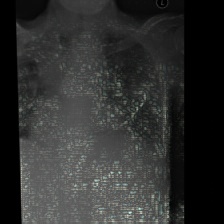}\\
\includegraphics[width=0.30\textwidth]{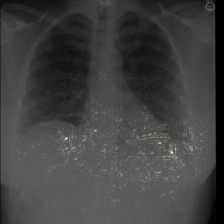}&\includegraphics[width=0.30\textwidth]{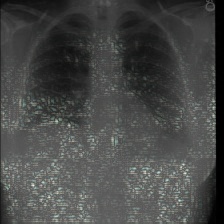}&\includegraphics[width=0.30\textwidth]{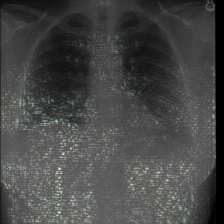}\\
\includegraphics[width=0.30\textwidth]{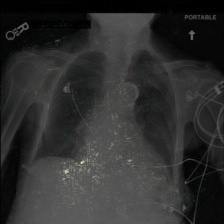}&\includegraphics[width=0.30\textwidth]{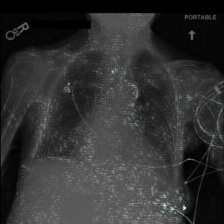}&\includegraphics[width=0.30\textwidth]{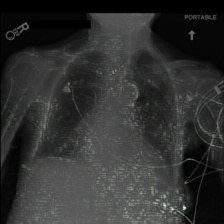}\\
\includegraphics[width=0.30\textwidth]{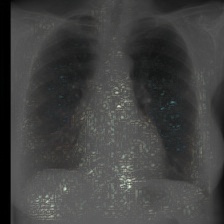}&\includegraphics[width=0.30\textwidth]{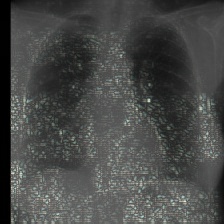}&\includegraphics[width=0.30\textwidth]{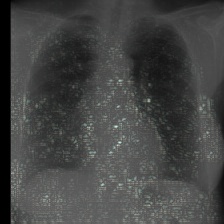}\\
        (a)&(b)&(c)
    \end{tabular}
    \caption{(a) The heatmap of SHAP values overlayed on a genuine sample from the MIMIC-CXR dataset, (b) The heatmap of SHAP values overlayed on the same image after being perturbed via PGD, (c) The heatmap of SHAP values overlayed on the same image after being perturbed by C\&W.}
\end{figure*}

\begin{figure*}
    \centering
    \begin{tabular}{ccc}
\includegraphics[width=0.30\textwidth]{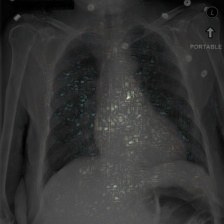}&\includegraphics[width=0.30\textwidth]{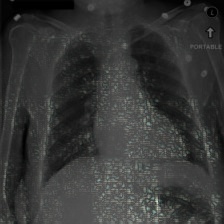}&\includegraphics[width=0.30\textwidth]{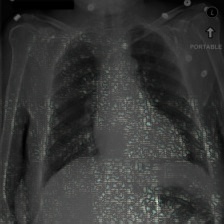}\\
\includegraphics[width=0.30\textwidth]{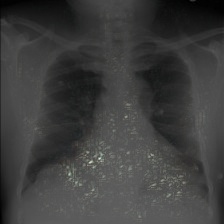}&\includegraphics[width=0.30\textwidth]{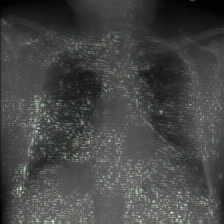}&\includegraphics[width=0.30\textwidth]{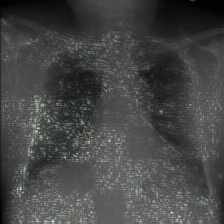}\\
\includegraphics[width=0.30\textwidth]{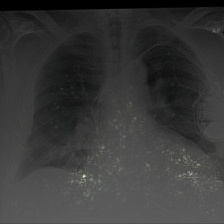}&\includegraphics[width=0.30\textwidth]{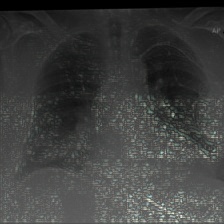}&\includegraphics[width=0.30\textwidth]{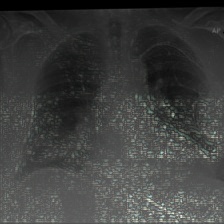}\\
\includegraphics[width=0.30\textwidth]{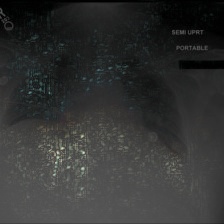}&\includegraphics[width=0.30\textwidth]{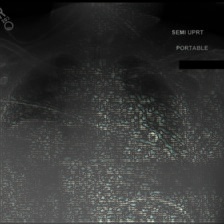}&\includegraphics[width=0.30\textwidth]{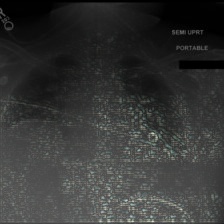}\\
        (a)&(b)&(c)
    \end{tabular}
    \caption{(a) The heatmap of SHAP values overlayed on a genuine sample from the MIMIC-CXR dataset, (b) The heatmap of SHAP values overlayed on the same image after being perturbed via PGD, (c) The heatmap of SHAP values overlayed on the same image after being perturbed by C\&W.}
    \label{fig:heatmapSHAPsupp}
\end{figure*}

\end{document}